\documentclass[letterpaper, 10 pt, conference]{ieeeconf}
\IEEEoverridecommandlockouts  %This command is only needed if you want to use the \thanks command

\usepackage{color}
\usepackage{array}

\usepackage{graphicx} % for pdf, bitmapped graphics files
\usepackage[hidelinks]{hyperref}

\usepackage{mathptmx} % assumes new font selection scheme installed
\usepackage{times} % assumes new font selection scheme installed
\usepackage{amsmath} % assumes amsmath package installed
\usepackage{amssymb}  % as0sumes amsmath package installed
\usepackage{color}
\usepackage{bm}
\usepackage{rotating}
\usepackage{subfigure}
\usepackage{tabularx}
\usepackage{colortbl}
\usepackage{hhline}
\usepackage{multirow}
\usepackage{verbatim}
\usepackage{cite}
\usepackage{siunitx}
\usepackage{graphicx}
\usepackage{duckuments}
\usepackage{soul,xcolor}
\setstcolor{red}
\usepackage[ruled, lined, linesnumbered, commentsnumbered, longend]{algorithm2e}
\usepackage[capitalise,noabbrev]{cleveref}
\usepackage{amsfonts}
\usepackage{siunitx}
\usepackage{booktabs}
\usepackage{gensymb}

\usepackage{multirow}
\usepackage[flushleft]{threeparttable}
\usepackage[acronym]{glossaries}
\glsdisablehyper

\newacronym{dof}{DoF}{Degree of Freedoms}
\newacronym{tof}{ToF}{Time of Flight}
\newacronym{hri}{HRI}{Human Robot Interaction}
\newacronym{fov}{FoV}{Field of View}
\newacronym{slam}{SLAM}{Simultaneous Localization and Mapping}
\newacronym{pf}{PF}{Particle Filter}
\newacronym{pc}{PC}{Point Cloud}
\newacronym{mcl}{MCL}{Monte Carlo Localization}
\newacronym{icp}{ICP}{Iterative Closest Point}
\newacronym{mocap}{MoCap}{Motion Capture}

\begin{document}
	
\title{\LARGE \bf
    Soft Robot Localization Using Distributed Miniaturized Time-of-Flight Sensors }

% Uncomment for IEEE template %%%%%%%%%%%%%%%%%%%%%%%%%%%%%%%%%%%%%%%%%%%%%%%%%%%%%%%

\author{\hspace{0.2in} Giammarco Caroleo$^{*}$, Alessandro Albini, Perla Maiolino
\thanks{The authors are with the Oxford Robotics Institute, University of Oxford, UK.}
\thanks{$^{*}$Corresponding author. Please, contact at the email address \texttt{giammarco@robots.ox.ac.uk}}
% <-this % stops a space
\thanks{This work was supported by the SESTOSENSO project (HORIZON EUROPE Research and Innovation Actions under GA number 101070310).}
}

\maketitle

%\linenumbers
%starts from a new page
\newpage
\begin{abstract}
Thanks to their compliance and adaptability, soft robots can be deployed to perform tasks in constrained or complex environments. 
In these scenarios, spatial awareness of the surroundings and the ability to localize the robot within the environment represent key aspects.
While state-of-the-art localization techniques are well-explored in autonomous vehicles and walking robots, they rely on data retrieved with lidar or depth sensors which are bulky and thus difficult to integrate into small soft robots.
Recent developments in miniaturized \gls{tof} sensors show promise as a small and lightweight alternative to bulky sensors. These sensors can be potentially distributed on the soft robot body, providing multi-point depth data of the surroundings.
However, the small spatial resolution and the noisy measurements pose a challenge to the success of state-of-the-art localization algorithms, which are generally applied to much denser and more reliable measurements.

In this paper, we enforce distributed VL53L5CX \gls{tof} sensors, mount them on the tip of a soft robot, and investigate their usage for self-localization tasks. Experimental results show that
the soft robot can effectively be localized with respect to a known map, with an error comparable to the uncertainty on the measures provided by the miniaturized ToF sensors.

\end{abstract}

\section{Introduction}\label{sec:intro}

\glsresetall

The compliance and adaptability of soft robots make them a valuable option to be deployed to perform tasks in constrained environments. Unlike traditional rigid robots, soft robots can deform into tight spaces, and navigate complex and cluttered surroundings safely~\cite{arezzo2017total, greer2018obstacle,wang2018toward, luong2019eversion, karimi_2023}. 
However, for these robots to operate effectively, they must also be localized with respect to the environment they are exploring~\cite{sorensen2021,rosi_2022}.
Therefore, self-localization or \gls{slam} is essential to achieve self-aware soft robots that can adapt their movements appropriately. 

Localization and \gls{slam} have been largely investigated for autonomous driving and walking robots~\cite{gouda_2013,rybczak2024,kazerouni2022,macario2022}. These systems are usually equipped with lidars or high-resolution RGB-D cameras to support algorithms performing mapping and localization.
Despite their reliability, these sensors cannot be easily integrated into small soft robots since they are heavy and bulky. Although~\cite{sorensen2021} has used RGB-D cameras to perform \gls{slam}, the soft robot was considerably bigger than a classical RGB-D camera. % into small soft robots such as \cite{}. 
An alternative solution more suitable for smaller soft robots is represented by miniaturized small RGB cameras. In~\cite{rosi_2022}, this solution has proven to be effective in estimating the posture of a soft robot through \gls{slam} algorithms. 
However, for soft robots, a single camera is typically integrated at the tip pointing downward \cite{rosi_2022, diodato2018soft,albeladi2022hybrid, arezzo2017total, kim2021origami}, thus limiting the \gls{fov}. A possible solution to increase spatial awareness consists of distributing sensors on the robot body. However, cameras cannot easily scale in number. The integration becomes difficult, not to mention the high bandwidth required to transmit and process data from multiple cameras. 
In this respect, the work in \cite{karimi_2023} equipped the soft robot using single-beam distance sensors, which are easier to distribute with respect to cameras. The work showed the possibility of incrementally reconstructing a point cloud of a 2D maze by acquiring samples from the distance sensors over time. These sensors, along with Inertial Measurement Units, were used to perform a \gls{slam} task in a maze. 

\begin{figure}[t!]
	\centering
	\includegraphics[width=0.98\columnwidth]{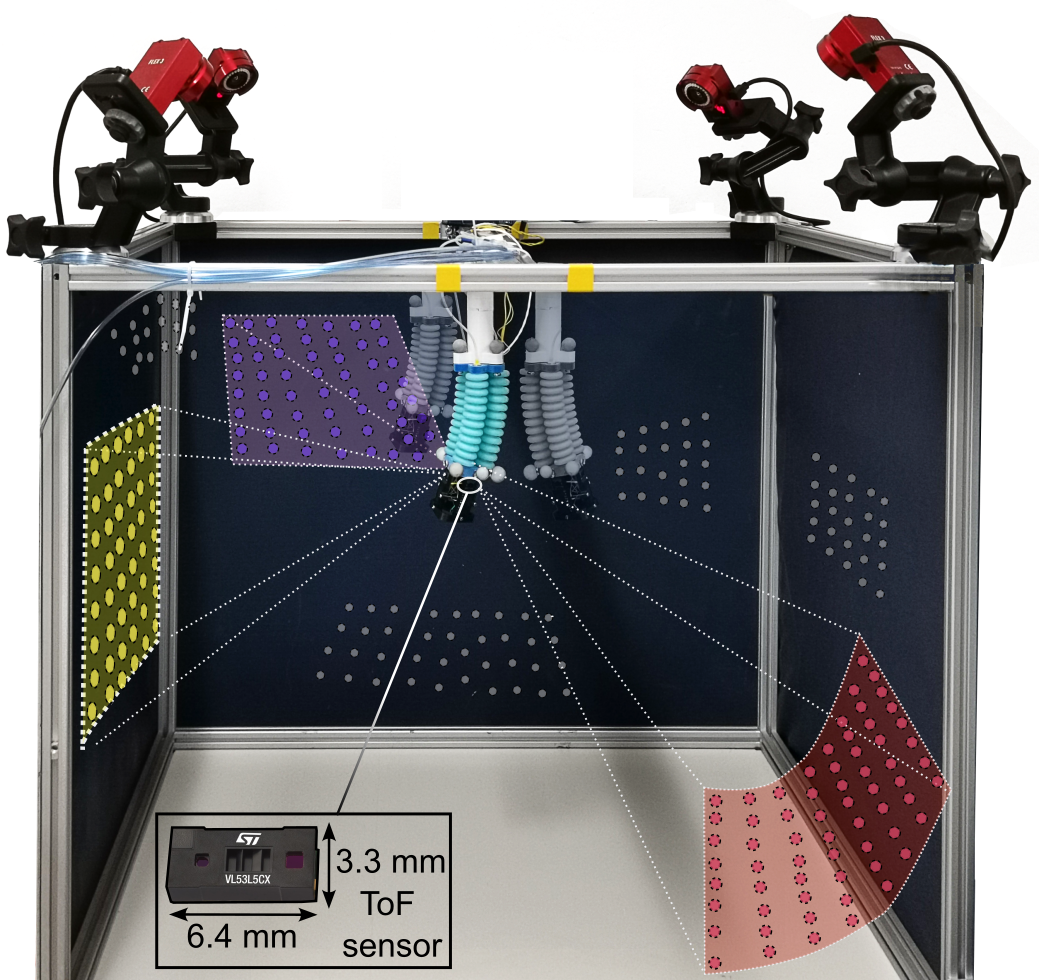}
	\caption{Soft robot estimating its pose with respect to the environment using a Particle Filter based on measurements collected with miniaturized ToF sensors. The faded robots represent the poses associated with different particles. For each sensor, the measured 8×8 depth map is converted into a point cloud and exploited for localization. The colored points belong to the point cloud associated with the current pose, while the shaded gray dots are the ones previously collected. The robot is successfully localized by using the coarse information retrieved with the ToF sensors.}
 % An image of the adopted sensor and its dimensions are also reported. 
	\label{fig:intro}
\end{figure}

Beyond the single-measurement distance sensors used in~\cite{karimi_2023}, miniaturized lidars based on \gls{tof} technology have been recently introduced. The measurement provided by this typology of \gls{tof} sensor consists of a low resolution depth map, which can be converted into a small point cloud representing the surrounding environment falling into the sensor's \gls{fov}.  
These sensors have generated significant interest among researchers since they are cheap, lightweight, small, and capable of multi-point sensing at medium range (typically up to \SI{4}{\meter})~\cite{hughes2018robotic, tsuji2021sensor,mu2024towards}.
So far, their usage in robotics has been predominantly related to collision avoidance, pre-touch sensing, in-hand manipulation, object detection, and gesture recognition, when equipped on rigid robots~\cite{ding2019proximity, ding2020collision, caroleo2024proxy, al2020towards, yang2017pre, sasaki2018robotic, koyama2018high, ruget2022pixels2pose}.

We argue that their compact form factor makes them suitable for distributed integration into the robot body, thereby enhancing its spatial awareness. However, it is worth noting that despite their advantages in terms of size compared to lidar or RGB-D cameras, they are significantly less reliable. In particular, miniaturized \gls{tof} presents the following drawbacks:
\begin{itemize}
    \item low spatial resolution - recent model can provide at most an 8x8 depth map;
    \item pyramidal \gls{fov} - the spatial resolution of the obtained point cloud is increasingly coarse depending on the distance between the robot and the environment;
    \item noise and uncertainties on measurements depend on environmental conditions and the distance with respect to the target. 
\end{itemize}
Performing self-localization or \gls{slam} is challenging with this type of sensing system. Indeed, state-of-the-art algorithms developed for these tasks are generally applied to much denser point clouds and more reliable measurements. 

In this paper, we want to investigate whether miniaturized \gls{tof} sensors, equipped on a soft robot, can be effectively used to perform self-localization with respect to a known map, using state-of-the-art algorithms.
Specifically, to estimate the robot's pose, we considered using a \gls{pf} algorithm whose particle weights are updated by considering the \gls{icp} score. This computes how well the point cloud reconstructed with \gls{tof} sensors aligns with the map when the transformation devised by the given particles is adopted.  

\cref{fig:intro} shows the miniaturized \gls{tof} equipped at the tip of the robot, illustrating their \gls{fov}. The points colored in red, yellow, and purple show the current measurements of the sensors, while the gray, the point cloud of the environment constructed by merging \gls{tof} measurements at different time instants.

The paper is structured as follows. 
\cref{sec:methodology} formally introduces the localization problem using distributed \gls{tof} sensors. The experimental setup and the validation experiments are described in~\cref{sec:validation}. Results and discussion are reported in~\cref{sec:results}. Conclusion follows.

\section{Localization based on Distributed ToF Sensors Measurements}\label{sec:methodology}

\subsection{Environment Reconstruction using Distributed ToF Measurements}

In this paper we consider a soft robot equipped with a set of \gls{tof} sensors  distributed on its body. The configuration of the soft robot is assumed to be measurable or at least estimated, while the pose of its base with respect to the environment is unknown. 
As shown in \cref{fig:intro}, the single sensor has a pyramidal \gls{fov}, and its output consists of a low-resolution depth map, where each measurement corresponds to the distance between the origin of the sensor and the portion of environment falling within the \gls{fov}.
Assuming that we have: (i) the intrinsic calibration of the sensors from datasheet specifications \cite{st_vl53l5cx_datasheet}, and (ii) their position with respect to a common reference frame (i.e., the robot base), it is possible to convert all the depth maps collected by the various sensors into a single point cloud $\bar{X}_k$. We refer to $\bar{X}_k$ as the \textit{point cloud sample}, where $k$ is the index representing the $k$-th sample.
By moving the robot into different positions, different parts of the environment will fall into the \gls{fov} of the sensors. Therefore, by collecting and merging samples at different robot configurations it is possible to incrementally build a point cloud representing the environment surrounding the robot.
This can be defined as  $X_k = \bar{X}_k \cup \bar{X}_{k-1}$, with $X_0$ being an empty point cloud set.

\subsection{Soft Robot Localization with PF}

Assuming the map of the environment $\hat{X}$ to be known in the form of a point cloud, the goal is to find an estimate $\hat{\mathbf{x}}\in SE(3)$ of the true pose of the robot base $\mathbf{x} \in SE(3)$ with respect to $\hat{X}$ from the knowledge of the measurements $\bar{X}_k$.
In this respect, we exploit the use of a \gls{pf} to estimate the pose of the robot $\hat{x}_k \in SE(3)$ at each $k$-th collected sample. A set of $M$ particles $P_k=\lbrace \mathbf{p}_k^1, \dots, \mathbf{p}_k^M \rbrace$ is used to describe the pose of the robot base with respect to the environment in the Cartesian space, with $\mathbf{p}_k^i \in SE(3)$. For the given $\bar{X}_k$, the particle weights are updated using the ICP registration algorithm~\cite{besl1992method} initialization step. More specifically, at the $k$-th iteration, for each devised pose $\mathbf{p}_k^i$, the $\bar{X}_k$ is transformed accordingly from the sensor reference frame to the robot base; then, the registration score  $w_k^i$ is computed with the initialization step of the ICP algorithm by matching $\bar{X}_k$ and $\hat{X}$, with $w_k^i \in [0,1]$. The resulting scores associated with the particles become crucial in the resampling phase in which new particles are devised, i.e., for the $(k+1)$-th step, particles are drawn with higher density in the neighborhood of the best matching poses of the $k$-th step by using the normal distribution. For $k=0$, particles are drawn from a Gaussian distribution while the associated weights are given by a uniform density distribution. 
The localization is accomplished by setting in advance the number $N$ of samples that need to be considered. For $k=N$, it is imposed $\hat{\mathbf{x}}= \frac{1}{M}\sum_{i=1}^{M}w_N^i\mathbf{p}_N^i$.

\section{Experiments Description}
\label{sec:validation}

\subsection{Experimental Setup}
\label{sec:setup}

\begin{figure}[]
	\centering
	\includegraphics[width=0.98\columnwidth]{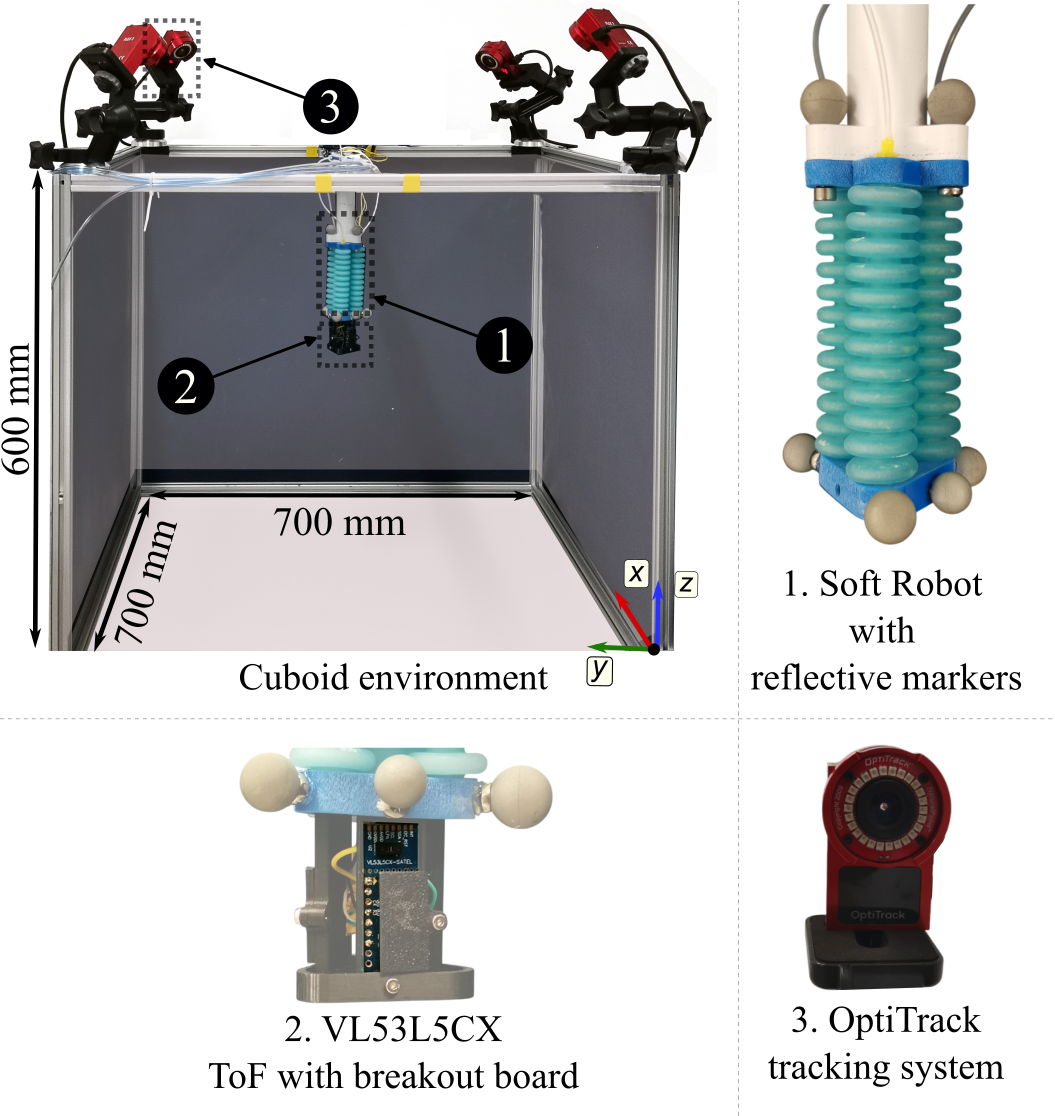}
	\caption{The experimental setup is shown. It consists of the cuboid environment (0.7x0.7x0.6 m) considered for the localization task (top left corner), the soft robot with reflective markers rigidly mounted on it (1), three VL53L5CX ToF sensors mounted on the robot tip (2), and the OptiTrack tracking system (3) with four cameras recording the reflective markers poses which constitute the ground truth for the pose estimation.}
	\label{fig:setup}
\end{figure}

We validated the proposed approach with the experimental setup showed in \cref{fig:setup}. 
The soft robot consists of a three-chambers bellow actuator where three \gls{tof} sensors have been integrated into the tip. The design of the actuator stems from the one adopted in ~\cite{veronese_robosoft} with the main difference being a central empty channel that accommodates the \gls{tof} wires. The multi-material 3D printer Stratasys J735\textsuperscript{\textregistered} (Stratasys Ltd, USA) was utilized to print in one batch both the soft bellows and the rigid edges used to connect the sensors support and the reflective markers. For the soft bellows, a Shore hardness of $40$A was achieved with a digital material blend of VeroCyan\textsuperscript{\textregistered} (Stratasys Ltd, USA) and the
rubber-like soft material Agilus30\textsuperscript{\textregistered} (Stratasys Ltd, USA). The support material required for the print was chemically dissolved in a tank by using a solution of \SI{0.02}{kg/L} Sodium Hydroxide and \SI{0.01}{kg/L}
Sodium Metasilicate. The printed actuator has an overall length of \SI{10}{cm}, while the diameter of its bellows is \SI{2.4}{cm}.

The specific range sensors used in this study are the VL53L5CX from ST MicroElectronics, featuring a pyramidal \gls{fov} of \SI{65}{\degree}. These are mounted through a rigid support connected to the tip of the robot, and their relative pose is given by the CAD model of the support. 
Each \gls{tof} sensor weighs approximately \SI{2}{\gram}. The majority of the weight of this setup is given by the breakout board and the prototyping electronics weighing around \SI{10}{\gram}. Sensors are connected with a microcontroller through an I2C bus. Each \gls{tof} sensor provides an $8 \times 8$ depth map at \SI{15}{\hertz} rate that can be converted into a point cloud as described in \cref{sec:methodology}.  

In order to create the point cloud of the environment $X_k$ which is updated with new samples, the relative configuration of the soft robot in space must be known. In this paper, we estimated the position of the robot's tip with respect to the robot base by training a simple $k$-NN regressor mapping the actuation pressure with the corresponding tip pose, measured by a \gls{mocap} system, as shown in~\cite{ouyang2022modular}. In this work, we used an OptiTrack with four cameras to precisely retrieve the position of the reflective markers attached to the tip. The position of the cameras with respect to the environment was such that the markers were clearly visible at all times throughout the experiments.

\subsection{Data Collection}
We performed the data collection described in the following to both: (i) train the $k$-NN model and (ii) collect point cloud samples.
The robot was actuated with increasing pressures ranging from 3 to 18 kPa, with increments of 1 kPa. We started inflating one chamber at a time and then actuated two chambers simultaneously using every combination of pressure values. This led to 807 different robot postures. For each commanded pressure combination, we waited \SI{4}{\second} to reach the steady state and then collected both the position of the tip, to train the $k$-NN model, and the \gls{tof} measurements, to be converted into point clouds samples.

The dataset composed of pressure sequence and tip poses measured by the OptiTrack was split into two parts, $80\%$ of the sample was used for training and the remaining $20\%$ for validating the model. The model is trained to predict the robot pose for a combination of pressures fed to the actuators. Similarly to~\cite{ouyang2022modular}, to devise the best-fitting number of neighbors for the model, the training set was also used in k-fold cross-validation~\cite{kohavi1995study}, with $k_{fold} = 5$. For the collected data, we found $k=6$ to be the best option with mean squared errors $MSE_t = 10^{-4}$ m on the training set and $MSE_v = 3\times10^{-4}$ m on the validation set. As will be extensively shown in~\cref{sec:results}, the high performance of the trained model allows for a seamless substitution of the OptiTrack tracking system since the robot's pose can be predicted with accuracy for a given pressure sequence fed to the actuators. 

For what concerns the collected point cloud samples, as explained in the next Section, various subsets of the 807 samples are randomly picked to generate the point cloud of the environment $X_k$ that is used for the localization task.

\subsection{Validation}

The proposed approach was validated in a localization task. The robot is rigidly mounted to the frame of the cuboid with a 3D printed support which constrains the robot's base to be parallel to the $x$-$y$ plane (see~\cref{fig:setup})
at a fixed, known distance. Due to this, the localization task simplifies to the estimation of the pose $\hat{\mathbf{x}}\in SE(2)$ as we consider uncertainty on the position in the horizontal plane and on the rotation around the $z$-axis. Formally, the goal is to estimate  $\hat{\mathbf{x}} = \lbrace \hat{x}, \hat{y}, \hat{\gamma} \rbrace$  with respect to the aluminum frame consisting of a cuboid with two open faces as can be seen in the figure.
A model of this environment was created with CAD software, from the knowledge of the frame's size, and exported as a point cloud consisting of 2000 points. 

The localization was performed with a \gls{pf} composed of 1000 particles. 
To evaluate how many point cloud samples are required to localize the robot we ran \gls{pf} using point clouds $X_k$ created with a variable number of samples. In particular, we evaluated the localization accuracy for point cloud composed of $k =[1, \dots, 10 ]$ samples, randomly picked from the 807 available. In addition to this, for a given number of samples, we performed the localization in 50 trials by randomly drawing diverse point clouds. 

Each point cloud has at most $64\times3 = 192$ points and is cropped to remove those points that belong to the environment outside the cuboid. The $k$-th sample is transformed with respect to the robot base using the $k$-NN model of the arm. Prior to merging $\bar{X}_k$ into $X_k$, in a similar fashion to what is proposed in~\cite{vizzo2023kiss}, the point clouds are downsampled using a voxel grid with a voxel size of \SI{0.05}{m}.

Particles were initialized as follows. The first two components of the particle, representing the translation on the plane were initialized with random numbers picked from a distribution centered on the true position of the robot base with a deviation varying in a range of \SI{0.2}{\meter}. Similarly, the third component of the particle, representing the in-plane angle, was initialized with a random distribution centered in the true angle with a deviation varying in a range of \SI{20}{\degree}.
For the update of the $i$-th particle weight $w_k^i$, the point-to-point ICP variant is adopted with considering the downsampled point cloud $X_k$, the map $\hat{X}$, and the transform devised by the particle $\mathbf{p}_k^i$.

It is also worth noting that, in general, for soft robots, the estimation of the robot configuration is affected by uncertainties~\cite{della2023model}. In this scenario, errors related to the tip's position are propagated to the corresponding point cloud, thus affecting the reconstruction of the environment and, as a consequence, the localization.
In this respect, we wanted to investigate how much the pose estimation performed with the $k$-NN model affects the localization accuracy. Therefore, we replicated the same experiments discussed above by generating the point clouds using measured robot tip position obtained with the \gls{mocap} system showed in~\cref{fig:setup}, whose precision is in the order of $10^{-4}$m.
 
\section{Results and Discussion}\label{sec:results}

\begin{figure}[b!]
	\centering
	\includegraphics[width=0.98\columnwidth]{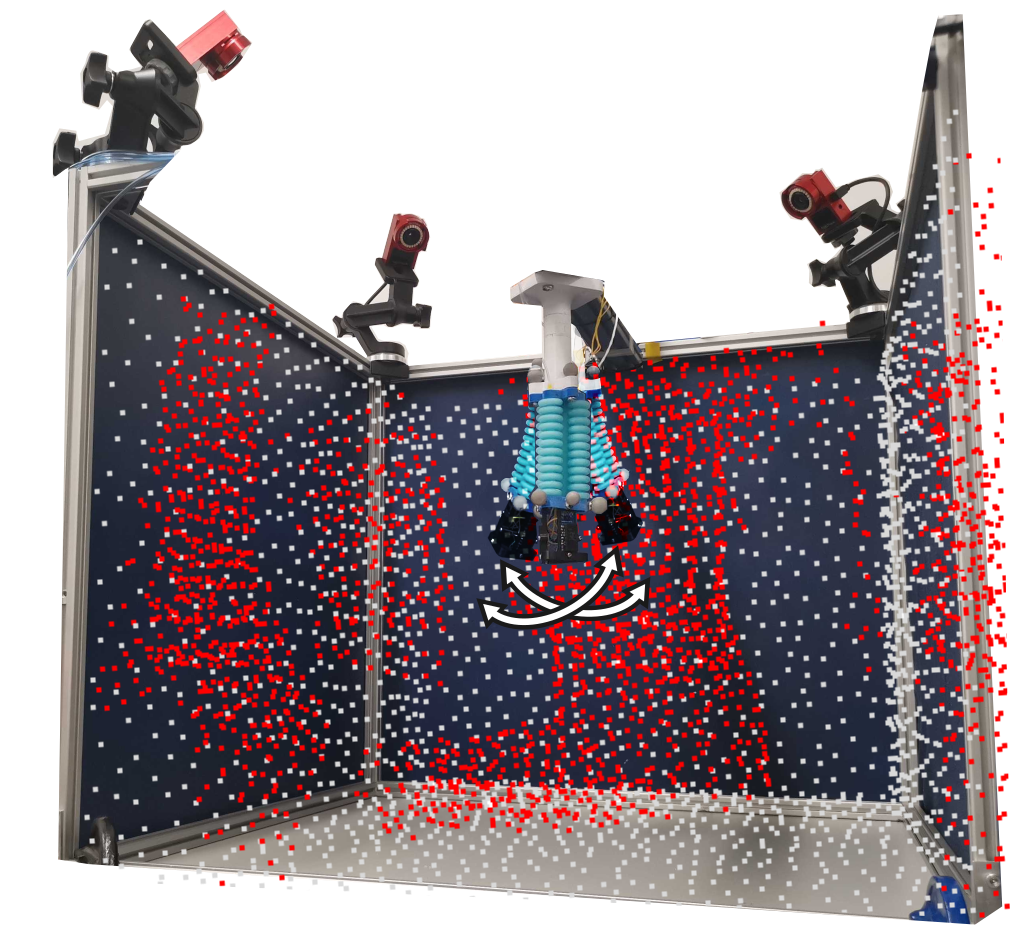}
	\caption{The model point cloud $\hat{X}$ is shown with gray dots and over-imposed on the real cuboid used for the localization task. The red point cloud $X_k$, was obtained by merging $k=50$ random point cloud samples recorded by the robot in different poses reached when commanded with $k$ pressure sequences. An exemplification of the soft robot in three different configurations is also shown. }
	\label{fig:reco}
\end{figure}

In order to show the quality of the point clouds obtained with these \gls{tof} sensors, we reported in \cref{fig:reco} a point cloud $X_k$ collected with $k = 50$ samples acquired with the robot in different poses. The figure also shows the ground truth map of the environment. As visible, measurements of the sensors are noisy, and despite the simple geometry of the surroundings, in some areas, it is not possible to reconstruct precisely the local shape. As an example, sharp features, such as corners, appear to be smoothed in the collected point clouds. This is due to the fact that ToF measurements are affected by environmental factors such as the multiple-ways reflections that can be induced by sharp edges and the reflectance of the scene~\cite{gudmundsson2007environmental, may2009robust} which in our case consists of metallic parts as well. Furthermore, the noise on the output readings also depends on the distance of the sensor from the target and their respective orientation which further complicates the success of the task.

\begin{figure}[t]
	\centering
	\subfigure[][] {\label{fig:results_knn}\includegraphics[width=0.9\columnwidth]{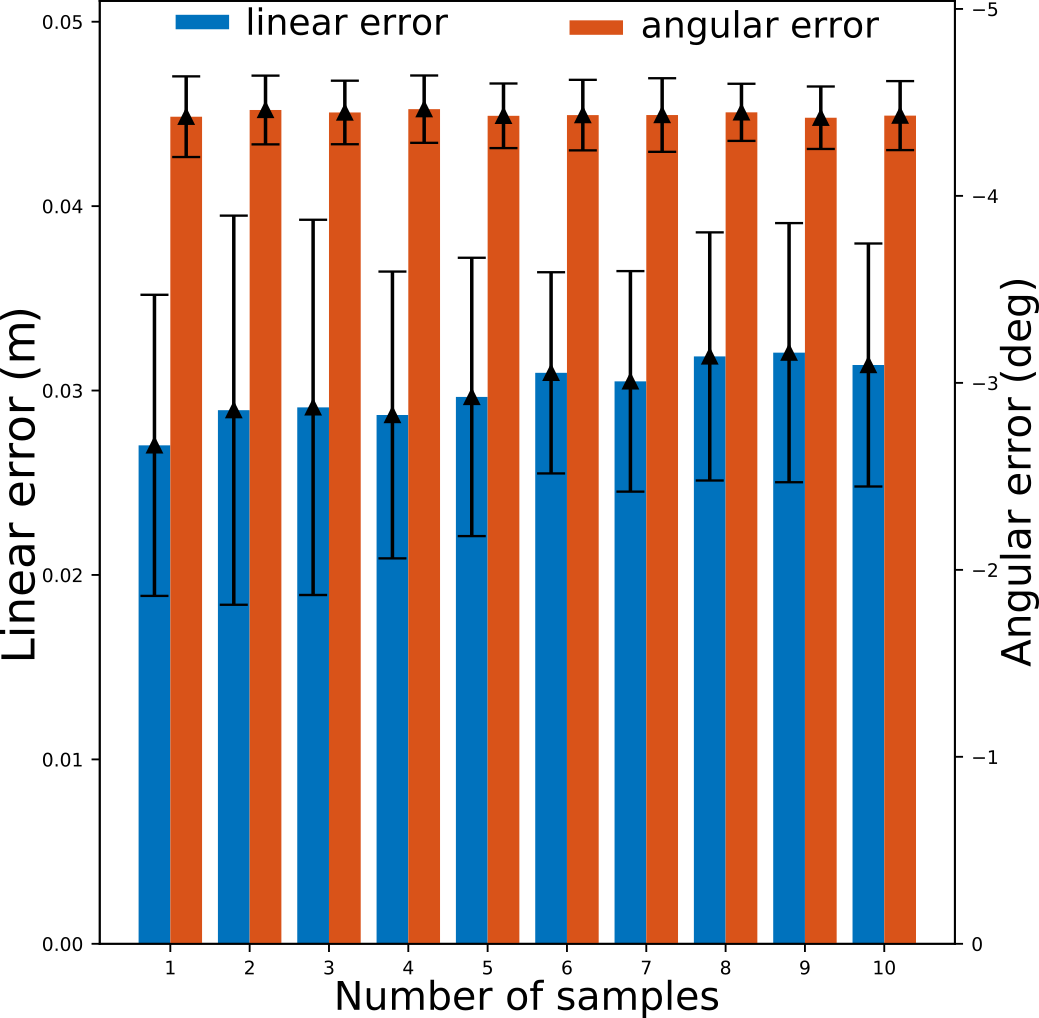}} 
	%%%%
	\subfigure[][]
	{\label{fig:results_opti}\includegraphics[width=0.9\columnwidth]{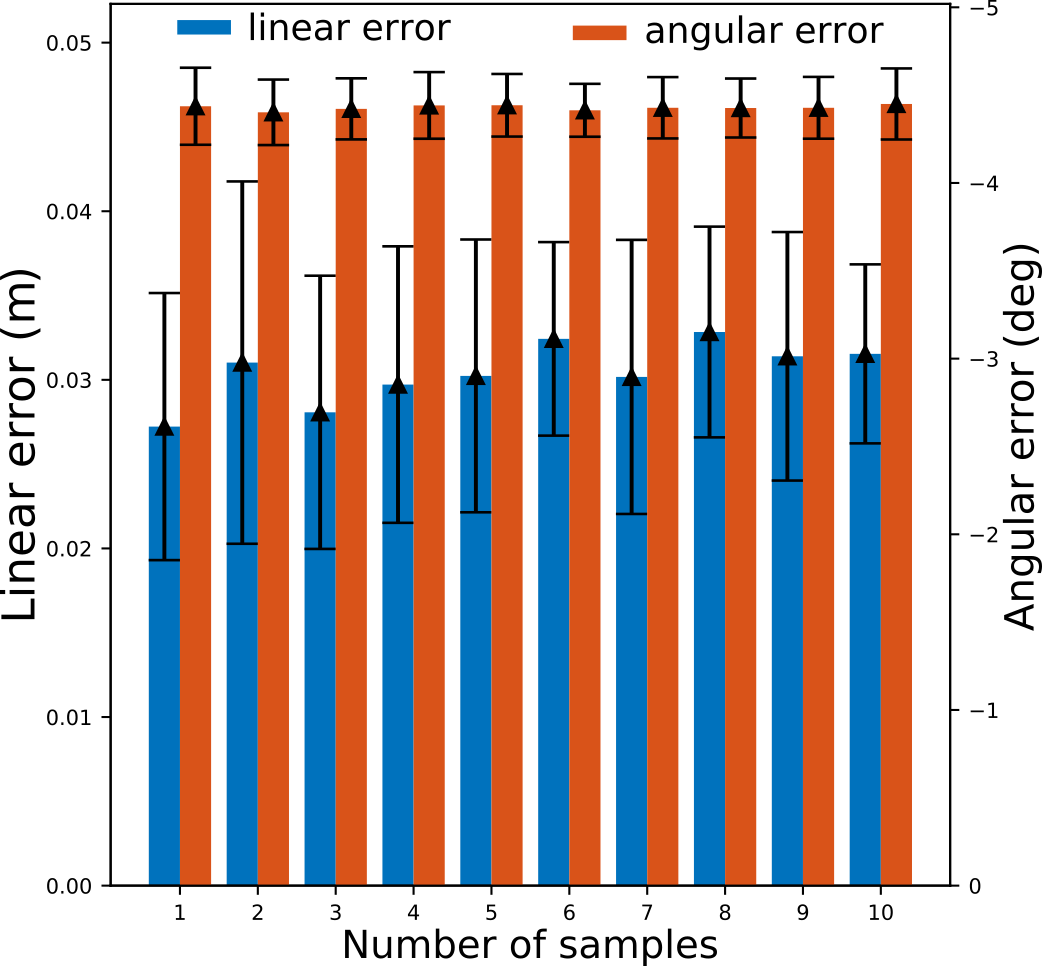}} 
	%%%%
	\caption{The bar plot shows the mean linear and rotational error along with their relative standard deviation ranges with a growing number of point cloud samples in two cases: (a) when point cloud samples are merged using the $k$-NN model for the transformation with respect to a common base frame; (b) when point cloud samples are merged using OptiTrack data for the transformation with respect to a common base frame. }
	\label{fig:results}
\end{figure}  

The performance on the localization task was evaluated by computing (i) the position error $e_x$ as the norm of the difference between the true and estimated position, and (ii) orientation error $e_{\gamma}$, i.e., the absolute value of the difference between the true and estimated in-plane angle of the robot base. 

Results are reported in~\cref{fig:results} for both point clouds merged using the tip poses estimated with the $k$-NN model (\cref{fig:results_knn}), and measured with the \gls{mocap} system (\cref{fig:results_opti}). 
The bars report the mean error across all trials for both position and orientation and the relative standard deviation.
It is visible from the plots that there is no significant difference between the two cases. This is further highlighted in~\cref{tab:table_errors}, where the average and the standard deviation across all the trials are given for the two considered approaches. Consequently, we can conclude that in this scenario, the localization accuracy is not affected by the approximate estimation of the robot tip position given by the $k$-NN model.

It is worth also noting that in both cases, increasing the point cloud samples does not lead to any improvements, as the localization error is always similar, with small oscillation in the average due to the noise of the measurements. We argue that this is because having a single actuator limits the motions of the robot, thus reducing the area explored. Additionally, the cuboidal shape of the environment does not have many discriminative features (only the corners), hence when the robot moves, it collects redundant information.

From \cref{fig:results} and \cref{tab:table_errors} it can be also seen that the mean error converges at approximately \SI{0.03}{\meter} and less than \SI{4.5}{\degree} for orientation. 
These results are in line with the information provided in the datasheet of the sensors used in this paper\cite{st_vl53l5cx_datasheet}. As previously mentioned in the introduction, one of the drawbacks of these sensors is the uncertainty of the measurement increasing with the distance between the sensor's origin and the environment. The size of the cuboid containing the robot is $0.7$ $\times$ $0.7$ $\times$ \SI{0.6}{\meter}. The robot is placed at the center of the frame, thus being distant approximately \SI{0.35}{\meter} from the cuboid vertical faces when not actuated. According to the sensor's datasheet, the uncertainty of the measurement in this range corresponds to 11\% of the real distance. %
Therefore, we should expect an uncertainty in the \gls{tof} measurements in the order of \SI{0.03}{\meter}, which is in line with our localization error.

For what concerns the distorted corners of the frame in some point clouds, we argue that the adopted updating policy for the particles is playing a major role in filtering them out. As a matter of fact, outliers are discarded in the ICP score; hence unreliable measures induced by multiple-ways reflections are not considered in the weighing process and do not hinder the success of the localization task.

\begin{table}[t!]
\label{tab:table_errors}
\caption{Mean and standard deviation of the linear and angular error when considering $k$-NN model and OptiTrack data to merge point cloud samples.}
\begin{tabular}{|cc|cc|}
\hline
\multicolumn{2}{|c|}{$k$-NN}                                       & \multicolumn{2}{c|}{OptiTrack}                                 \\ \hline
\multicolumn{1}{|c|}{$e_x$ (m)}           & $e_\gamma$ ($\degree$) & \multicolumn{1}{c|}{$e_x$ (m)}           & $e_\gamma$ ($\degree$) \\ \hline
\multicolumn{1}{|c|}{0.0300 ± 0.0076} & 4.438 ± 0.181            & \multicolumn{1}{c|}{0.0305 ± 0.0076} & 4.428 ± 0.182            \\ \hline
\end{tabular}
\label{tab:table_errors}
\end{table}

We argue that to improve the performance of the \gls{pf} a probabilistic model, such as the beam model \cite{thrun2002probabilistic}, may be adapted to work on these specific types of \gls{tof} sensors, thus modeling the uncertainty on the measurements.
In addition to this, it is also important to note that, although the environment has a simple shape, the localization can be challenging - a cuboid presents symmetries and large flat areas, making it hard to find distinguishing features that can be used to match the point clouds, especially when only a partial representation of the environment is available. Adding more links to the robot and additional \gls{tof} sensors can represent a solution to increase its range of motion and capture additional areas of the environment. Furthermore, another aspect worth investigating is to consider more complex environments. Indeed, they may be composed of more discriminative features that can make it easier to match the reconstructed point cloud with the known map of the environment. These aspects will be investigated in a future extension of the paper.

\section{Conclusion}\label{sec:conclusion}

In this paper, we evaluated the use of miniaturized \gls{tof} sensors to localize a soft robot with respect to a known environment.
Due to their reduced weight, the main advantage of these sensors is the possibility of distributing them into the soft robot to enhance its spatial awareness.
However, compared to traditional sensors commonly used for localization they present several drawbacks, such as significantly lower spatial resolution and reliability of the measurements.

Experimental results show that by applying state-of-the-art methods for localization, such as \gls{pf}, it is possible to localize the robot in a known map with an accuracy comparable to the uncertainty of the measurement provided by the sensors.

Future works will be dedicated to integrating probabilistic models to take into account the uncertainties provided by the sensor, thus possibly improving the localization accuracy, and 
assessing the performance of the soft robot localization in more complex environments.

\section*{Acknowledgment}\label{sec:acknowledgment}
The authors would like to acknowledge Yao Yao (University of Oxford) for the support provided with the manufacturing of the bellow actuator. 

{\small
\bibliographystyle{IEEEtran}
\bibliography{ref}
}

\end{document}